\documentclass[sigconf,nonacm,natbib=true]{acmart}

\usepackage{amsmath}
\usepackage{bbm}
\usepackage{subfig}
\usepackage{booktabs}
\usepackage{wrapfig}
\usepackage[export]{adjustbox}
\usepackage{subfig}
\usepackage[labelfont=bf, justification=justified]{caption}

\usepackage{booktabs}
\usepackage{multirow}
\usepackage[normalem]{ulem}
\useunder{\uline}{\ul}{}
\usepackage{amsmath,amsfonts,bm}

\def\eqref#1{equation~\ref{#1}}

\def\1{\bm{1}}

\def\rmX{{\mathbf{X}}}
\def\rmY{{\mathbf{Y}}}
\def\rmZ{{\mathbf{Z}}}

\def\mC{{\bm{C}}}

\def\mZ{{\bm{Z}}}

\DeclareMathAlphabet{\mathsfit}{\encodingdefault}{\sfdefault}{m}{sl}
\SetMathAlphabet{\mathsfit}{bold}{\encodingdefault}{\sfdefault}{bx}{n}

\def\gD{{\mathcal{D}}}

\def\gL{{\mathcal{L}}}

\newcommand{\E}{\mathbb{E}}

\newcommand{\TheName}{\texttt{ContRE}}

\AtBeginDocument{%
  \providecommand\BibTeX{{%
    \normalfont B\kern-0.5em{\scshape i\kern-0.25em b}\kern-0.8em\TeX}}}

\begin{document}
%
\title{Practical Assessment of Generalization Performance Robustness for Deep Networks via Contrastive Examples}

%
%

\author{Xuanyu Wu}
\authornote{Equal contribution.}
\authornote{Work done as intern at Baidu Research, Baidu Inc.}
\affiliation{%
 \institution{University of Pennsylvania, Philadelphia, USA}
 \country{}}
 
\author{Xuhong Li}
\authornotemark[1]
\affiliation{%
 \institution{Baidu Research, Baidu Inc., China}
 \country{}}

\author{Haoyi Xiong}
\authornote{Corresponding author.}
\affiliation{%
 \institution{Baidu Research, Baidu Inc., China}
 \country{}}
 
\author{Xiao Zhang}
\affiliation{%
 \institution{Tsinghua University, Beijing, China}
 \country{}}
 
\author{Siyu Huang}
\affiliation{%
 \institution{Nanyang Technological University, Singapore}
 \country{}}
 
\author{Dejing Dou}
\affiliation{%
 \institution{Baidu Research, Baidu Inc., China}
 \country{}}

\begin{abstract}

Training images with data transformations have been suggested as \textit{contrastive examples} to complement the testing set for generalization performance evaluation of deep neural networks (DNNs)~\cite{jiang2020neurips}. In this work, we propose a practical framework \TheName{}\footnote{In French, the word ``contre'' means ``against'' or ``versus''.} that uses {\ul Cont}rastive examples for DNN gene{\ul R}alization performance {\ul E}stimation. Specifically, \TheName{} follows the assumption in~\cite{chen2020simple,he2020momentum} that robust DNN models with good generalization performance are capable of extracting a consistent set of features and making consistent predictions from the same image under varying data transformations. Incorporating with a set of randomized strategies for well-designed data transformations over the training set, \TheName{} adopts classification errors and Fisher ratios on the generated contrastive examples to assess and analyze the generalization performance of DNN models in complement with a testing set. To show the effectiveness and efficiency of \TheName{}, extensive experiments have been done using various DNN models on three open source benchmark datasets with thorough ablation studies and applicability analyses. Our experiment results confirm that (1) behaviors of deep models on contrastive examples are strongly correlated to what on the testing set, and (2) \TheName{} is a robust measure of generalization performance complementing to the testing set in various settings.

\end{abstract}

\keywords{Contrastive Examples, Robustness, Generalization Performance}

%

\maketitle

\section{Introduction}

\begin{figure}[t]
    \centering
    \includegraphics[width=\linewidth]{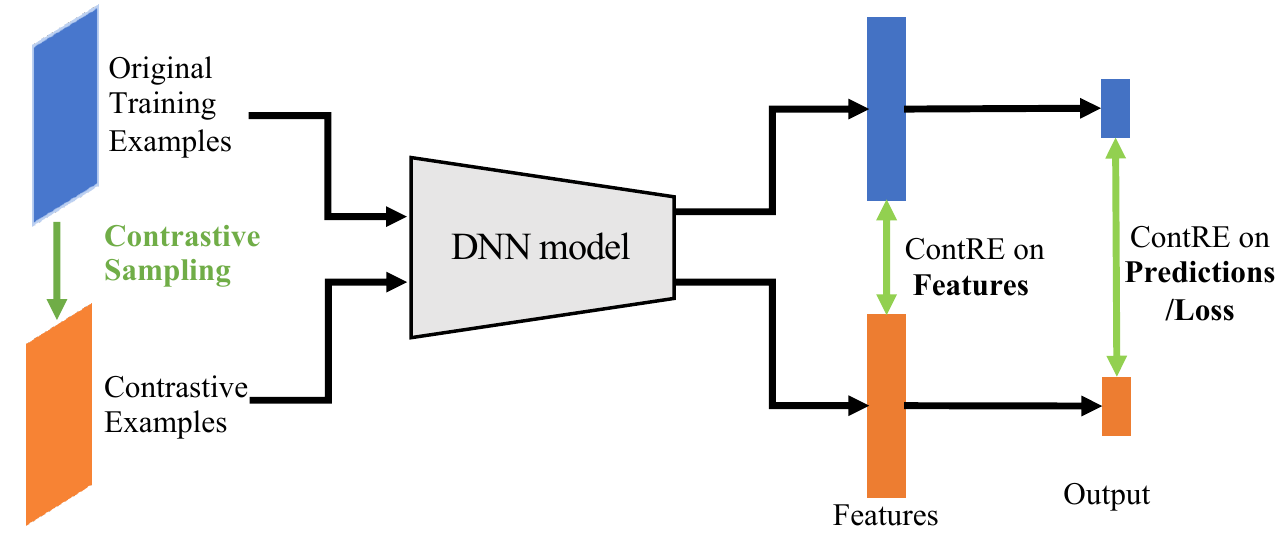}
    \caption{Illustration of the proposed framework \TheName{}. \TheName{} records and compares features and prediction results of the original images from the training set with the results of the contrastive examples. }
    \label{fig:pipeline}
\end{figure}

Deep neural networks (DNNs) have achieved remarkable performance in various domains such as computer vision, natural language processing, acoustic etc, but it is still challenging to assess the generalization performance of models in a robust way. 
Such robustness characterizes the ability to measure the performance of models on unseen data.
%
Given a DNN model trained with a sample set, a proxy to evaluate the generalization performance is to evaluate the model using samples in a disjoint testing set drawn from the same underlying data distribution.
We can then approximate the generalization error of the model by using the testing error, instead of coping with the intractable expectation of classification error over the data distributions (for both seen and unseen samples).

In addition to the testing set, many works nowadays have made lots of efforts on discovering intrinsic properties of deep models, such as weight norms, gradient norms, sharpness of the local minima etc.~\cite{Arora2018stronger,Neyshabur2017exploring,jiang2018predicting,jiang2020Fantastic}, that are correlated to the generalization performance. 
The use of these intrinsic properties as metrics could evaluate the generalization performance of a model, 
while avoiding ``overfitting'' to the validation/testing set in performance tuning~\cite{recht2019imagenet}. However, most of these metrics \textit{are vacuous on current deep learning tasks}~\cite{jiang2020Fantastic} and lead to poor predictability of generalization performance in some practical settings. 
%
%

Recently, a practical approach to evaluate generalization performance that exploits the training samples with image transformations, e.g., crops, shifts, rotations and color distortions, has been discussed in~\cite{jiang2020neurips} and suggested as a potential way to assess the generalization performance using such \textit{contrastive examples} (named after contrastive learning~\cite{chen2020simple,chen2020mocov2} for the transformed examples).
%
%
Though this approach could be loosely backed by some theoretical results~\cite{bousquet2002stability,deng2020toward}, the empirical use of contrastive examples for performance evaluation would be less effective when the similar data transformations have been used for training data augmentation to produce the model. 
This failure can be well explained by the high capacity of DNNs in memorization to overfit every seen sample~\cite{zhang2016understanding}. Thus, there needs a way to adapt the data transformation to arbitrary models for generalization performance measurements, even when the ways of training data augmentation are varying.



In this work, we propose \TheName{} that uses \underline{Cont}rastive examples for gene\underline{R}alization performance \underline{E}valuation.  Specifically, \TheName{} records features and prediction results of the original training set and what of the contrastive examples (see Figure~\ref{fig:pipeline} for an illustration).
For generalization performance evaluation, we follow the assumptions in~\cite{chen2020simple,chen2020mocov2} that a generalizable DNN model should make consistent prediction for the same image of different views transformed by varying strategies, as the model would extract an invariant set of features from the image. In this way, a model with higher/lower consistency in prediction results and extracted features under varying data transformations would be with better/worse generalization ability. 
Hereby, \TheName{} proposes to evaluate the generalization performance of DNN models using the classification accuracy (and Fisher ratio of feature vectors) on a set of contrastive examples. To generate effective contrastive examples, \TheName{} adopts RandAugment~\cite{cubuk2020randaugment}, to ensemble tens of commonly-used image transformations in a stochastic manner, where every original training sample would be transformed by a sequence of randomly picked-up operations.

In summary, this work makes contributions as follows:
\begin{itemize}
    \item 
    We propose a simple yet effective framework \TheName{} that generates contrastive examples using randomized data transformations on the training data and measures classification errors (and Fisher ratios) of DNN models on contrastive samples for the generalization performance evaluation purpose.
    
    \item We conduct extensive experiments using various DNN models on three benchmark datasets, i.e., CIFAR-10, CIFAR-100 and ImageNet, empirically demonstrating the strong correlations between the accuracy on contrastive examples and the one on the testing set. 
    
    \item Thorough analyses are provided: (1) By breaking down the data transformation strategies used in \TheName{}, we systematically analyze the effect of every single transformation technique and combination effects of all possible pairs; (2) We provide applicability analyses to confirm the versatility of \TheName{} working with arbitrary models trained with varying data augmentations; (3) We further investigate when and why the \TheName{} an effective way to estimate the generalization performance of DNN models.
    
\end{itemize}
To the best of our knowledge, this work is the first to practically assess the relevance of using contrastive examples for generalization performance of DNN models.
%



\section{Related Work}
In this section, we introduce the related work from the methodology and application perspectives.

\paragraph{Contrastive Learning and Contrastive Examples} The idea of contrastive learning is to pull closer the representations of different views of the same image (positive pair) and yet repel representations of different images (negative pair). Networks are then updated iteratively with positive and negative pairs through various contrastive learning objectives~\cite{he2020momentum,chen2020simple,chen2020big,chen2020mocov2,caron2020unsupervised}.
In a contrastive learning framework, a stochastic data transformation module of generating different correlated views of the same image is proved to be effective to representation learning.
Besides usages in contrastive representation learning, data augmentation has also been shown essential to improve the generalization of DNN models.
The trivial ones, such as random crop, random flip etc, are widely used for training large DNN models.
More recently, automated data augmentation policies including AutoAugment \cite{ekin2018auto} have been proposed to select data augmentation methods for boosting the model performance.
Rather than selecting data augmentation methods and compositions from a search space, RandAugment \cite{cubuk2020randaugment} proposes to randomly choose a number of augmentation distortions from a transformation candidate pool with a preset distortion magnitude.

In this work, we propose to measure the robustness of DNN models based on the (dis)similarity on features and final predictions between original images and contrastive examples, where we follow the RandAugment module to generate contrastive examples.

\paragraph{Generalization Performance and Robustness} Generalization performance measures the behaviors of models on unseen data. 
With the hold-out testing set as a proxy for measuring the generalization performance, the generalization gap is defined as the difference between the accuracies on the training set and the testing set. 
While previous works developed bounds for DNN models' generalization performance~\cite{Arora2018stronger, Neyshabur2017exploring}, some recent works predict generalization gap from the statistics of training data or trained model weights. Jiang et al. \cite{jiang2018predicting} approximate the margin in neural networks and train a linear model on margin statistics to predict the generalization gap of deep models. Yak et al. \cite{Yak2019TowardsTA} build upon the work by learning DNNs and RNNs in place of linear models. Instead of exploiting the margins, Unterthiner et al. \cite{thomas2019predicting} build predictors from model weights. Corneanu et al. \cite{corneanu2020computing} define DNNs on the topological space and estimate the gaps from the topology summaries.
Note that \TheName{} could be loosely guaranteed through data-dependent theoretical bounds~\cite{bousquet2002stability} and potentially related to the local elasticity~\cite{he2019local,deng2020toward} while the theoretical adaptation to DNN models should be more cautiously considered. 
We leave this as future work.

Another way of analyzing generalization performance is through complexity measures since a lower complexity often implies a smaller generalization gap~\cite{mcallester1999pac, keskar2016on, liang2017fisher, nagarajan2019generalization, neyshabur2015path, peter2017spectrally,jiang2020Fantastic}. The complexity is roughly defined as the measurement based on the DNN model's properties such as the norm of parameters and ``sharpness'' of the local minima. We refer interested readers to Jiang et al. \cite{jiang2020Fantastic}, where authors did extensive experiments to test the correlation between 40 complexity measures and deep learning models' generalization gaps, suggesting that PAC-Beyasian bounds \cite{mcallester1999pac} and sharpness measure \cite{keskar2016on} perform the best overall. 


Different from theoretical investigations and complexity measure of DNN models, \TheName{} measures the consistency of features and prediction results between contrastive examples and original images as generalization performance, which is data-dependent and practically works well. 



\section{Contrastive Sampling Framework for DNN Robustness Evaluation}

For leveraging large-scale unlabeled images, visual contrastive learning approaches \cite{oord2018representation,tian2019contrastive,henaff2020data,he2020momentum,chen2020simple} propose to train DNN models for solving the self-supervised contrastive prediction task that classifies whether each pair of images is similar or dissimilar \cite{hadsell2006dimensionality}.
Similar pairs are designed by a simple yet effective paradigm SimCLR \cite{chen2020simple}, that transforms original images through various random data augmentation approaches \cite{cubuk2020randaugment}, and then considers the pairs of augmented samples derived from the same images as similar pairs and others as dissimilar ones.
We call these generated samples as \textit{contrastive examples} and the process of generating contrastive examples as \textit{contrastive sampling}.
Contrastive sampling is crucial in the self-supervised training of a DNN model that yields general and discriminative representations.

In this paper, instead of aiming at representation learning, we propose to investigate the robustness and consistency behaviors of (supervised) trained DNN models through our proposed approach \TheName{}, the \underline{Cont}rastive sampling framework for DNN \underline{R}obustness and generalizability \underline{E}valuation.
Figure~\ref{fig:pipeline} shows the process of \TheName{}.
Specifically, \TheName{} first generates contrastive examples using RandAugment \cite{cubuk2020randaugment} and forwards generated samples into DNN models.
Given the intermediate features and predictions of DNN models with contrastive examples as input, \TheName{} compares the results with original samples as input.


In the rest of this section, we formally introduce the proposed framework \TheName{} and the evaluation metrics for the investigations and analyses, as well as some discussions about \TheName{}.

\subsection{Notations}

For convenience, we consider a DNN model as composition of the classifier $g$ and the feature extractor $f^l$, where $l$ indicates the layer index and is omitted for the last layer before the classifier.
From an underlying data-generating distribution $\gD$, an empirical dataset for training $\mZ$ and a disjoint testing set $\mZ_{test}$ are drawn.
Given a DNN model (with parameters $\bm{\theta}$) trained on $\mZ$, we can simply denote all the forwarding results of this DNN model, including intermediate features $f^l (\mZ; \bm{\theta})$, the estimates of the DNN model $g \circ f (\mZ; \bm{\theta})$, and the loss/accuracy on the training set with an objective function $\gL_{\bm{\theta}}(\mZ)$, as well as the generalization performance $\E_{z \sim \gD} \gL_{\bm{\theta}}(z)$.
In practice, the generalization performance is measured on the disjoint set $\gL_{\bm{\theta}}(\mZ_{test})$.
Moreover, a set of DNN models with different parameters are considered: $\{\bm{\theta}_1, \bm{\theta}_2, ...\}$.

\subsection{Contrastive Examples}

Given the notations, we introduce the proposed framework \TheName{}, which considers a number of widely-used data transformation approaches
to generate a different view of the original dataset as the contrastive sampling process.
We note the contrastive sampling strategy as $R$ and the obtained contrastive examples as $\mZ^s = R(\mZ) \sim R(\gD)$.
$R$ can be any data transformation method.
Besides the trivial ones, \TheName{} follows the strategy of RandAugment \cite{cubuk2020randaugment} for generating contrastive examples.

According to our preliminary experiments, any relevant image transformation method is effective to the generalization estimation if the transformed images have not been remembered by the model during training.
In reverse, the unseen transformed images can be always added as data augmentation techniques for improving the performance.
We could not easily identify the used transformation methods nor regulate the data augmentation if the access to the training process is not available.
To this end, \TheName{} adopts the strategy of RandAugment to increase the difficulty of fitting all the possible transformed images by the DNN models.

Given the contrastive examples, the primary objective of \TheName{} is to exploit the contrastive sampling strategy to analyze and evaluate the robustness of DNN models.
To achieve the goal, with contrastive examples or original images as input, \TheName{} thoroughly compares the behaviors of models on the prediction accuracy and intermediate features.
Formally, given $\mZ^s$, \TheName{} computes the intermediate features $f^l (\mZ; \bm{\theta})$ and $f^l (\mZ^s; \bm{\theta})$, prediction or loss $\gL_{\bm{\theta}}(\mZ)$ and $\gL_{\bm{\theta}}(\mZ^s)$, across different architectures and parameters $\{\bm{\theta}_1, \bm{\theta}_2, ...\}$.
Given these features and performances of various DNN models, \TheName{} measures their robustness and analyzes the consistency, with the metrics described below.

\subsection{Correlations and Fisher Ratio}

Despite the strong non-linearity, DNN models exhibit relatively similar/consistent behaviors with similar samples as input (except the devised adversarial attacks \cite{szegedy2013intriguing,goodfellow2014explaining,andrew2019adversarial,hamid2020on}).
For example, a large crop or a vertical flip of an image rarely changes the DNN estimates; while a rotation or a severe color distortion slightly disturbs its predictions.
Nevertheless, the level of robustness differs across models and this difference may be exploited to investigate the relation with the robustness and generalization performance.
We introduce two correlation metrics and one feature clustering quality metric that are used for experimental evaluations.

\subsubsection{Spearman's Rank Correlation Coefficient}
The proposed framework \TheName{} provides a good practical estimator of DNN models' generalization performance by measuring the loss/accuracy with contrastive examples as input.
Specifically, given $m$ DNN models, we measure the Spearman's rank correlation coefficient, noted as $r_s$ between two rank variables:
\begin{equation}
\label{eq:corr}
    r_s = \frac{\text{cov}(\rmX, \rmY)}{\sigma_\rmX \sigma_\rmY},
\end{equation}
where $\text{cov}(\rmX, \rmY)$ is the covariance, $\sigma_\rmX$ and $\sigma_\rmY$ are the standard deviations of the rank variables.
Specifically, the rank variables are converted from $\{\gL_{\bm{\theta}_i}(\mZ_{test})\}_{i=1,2,...,m}$ and $\{\gL_{\bm{\theta}_i}(\mZ^s)\}_{i=1,2,...,m}$.

We show in the experimental section that this strong correlation exists in various scenarios, across datasets, hyper-parameter settings, image transformation techniques, etc.


\subsubsection{Partial Rank Correlation}
Only using correlation as the measure for association between two variables can be misleading because their dependence may come from the associations of each with a third confounding variable. The problem can be solved by controlling the confounding variable via partial correlation. Let $\rmX$, $\rmY$ and $\rmZ$ be random variables taking real values, $r_{\rmX\rmY}$ be Spearman correlation between $\rmX$ and $\rmY$, then the partial rank correlation given control variable $\rmZ$ is:
\begin{equation}
    r_{\rmX\rmY,\rmZ} = \frac{r_{\rmX\rmY}-r_{\rmX\rmZ}r_{\rmY\rmZ}}{\sqrt{1-r_{\rmX\rmZ}^2}\sqrt{1-r_{\rmY\rmZ}^2}}. \label{eq:2}
\end{equation}
We use the partial correlation to eliminate the effect of training accuracy $\{\gL_{\bm{\theta}_i}(\mZ)\}_{i=1,2,...,m}$  while investigating \TheName's results. The details can be found in the experiment section.


\subsubsection{Fisher Ratio}
To verify the existence of significant correlations on feature level, \TheName{} also proposes to investigate the clustering quality of intermediate features. 
The degree of separation of features from other classes can be measured elegantly by the Fisher ratio. Given $g$ different classes, there are $N_i$ data points in class $\pi_i$, with class mean $\bar{x_i}=\frac{1}{N_i}\sum_{j=1}^{Ni}x_{i,j}$. 
The between class scatter matrix measures mean separation and the within class matrix measures class concentration by pooling the estimates of the covariance matrices of each class, which are defined as:
\begin{equation}
    \left\{
         \begin{array}{l}
         S_b = \sum_{i=1}^{g}N_i(\bar{x_i}-\bar{x})(\bar{x_i}-\bar{x})^T,  \\
         S_w = \sum_{i=1}^{g}\sum_{j=1}^{N_i}N_i(x_{i,j}-\bar{x_i})(x_{i,j}-\bar{x_i})^T.
         \end{array}
    \right.
\end{equation}
Then the Fisher ratio is:
\begin{equation}
    Fisher Ratio = Trace(S_w^{-1}S_b).
\end{equation}
Fisher ratio is a powerful tool for analyzing models' generalization capability right before the classifier. We compare features extracted from original samples and contrastive examples using Fisher ratio, i.e., $\{f^l (\mZ; \bm{\theta}_i)\}_{i=1,2,...,m}$ and $\{f^l (\mZ^s; \bm{\theta}_i)\}_{i=1,2,...,m}$, as part of explanations of why \TheName{} works well. The detailed findings are located in the experiment section.

\section{Experiments and Analyses}

\begin{figure*}[t]

\subfloat{%
  \includegraphics[clip,width=0.95\textwidth]{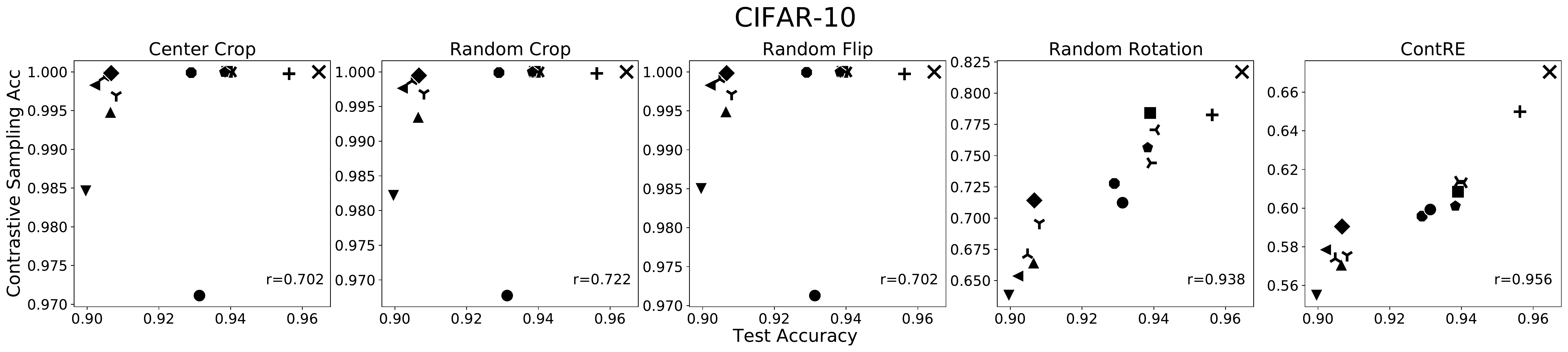}%
}

\subfloat{%
  \includegraphics[clip,width=0.95\textwidth]{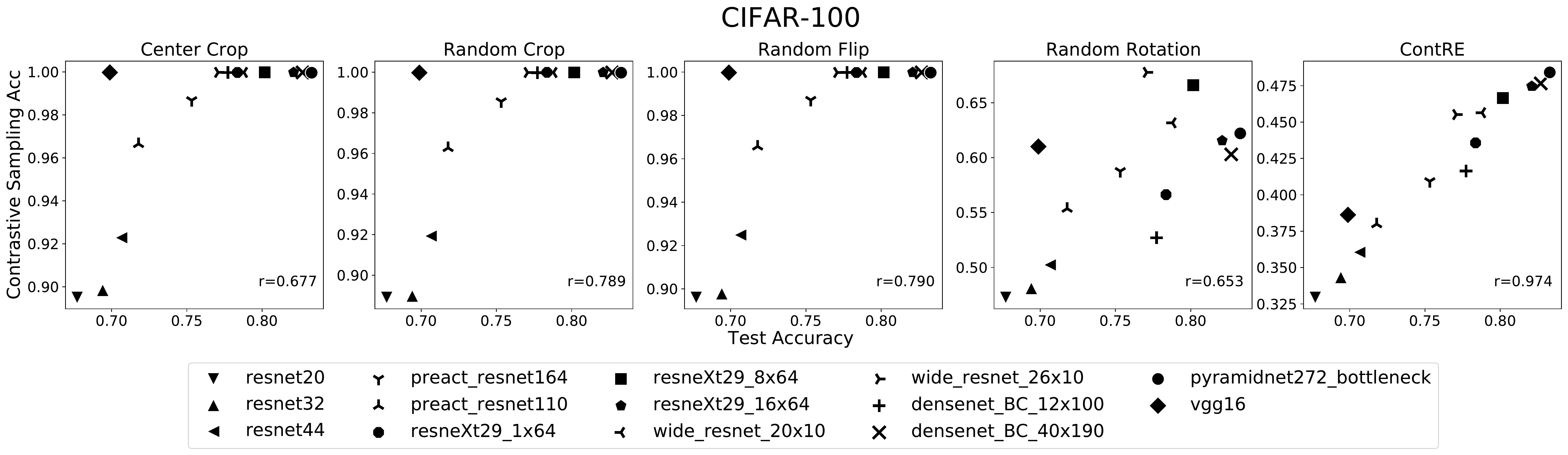}%
}

\subfloat{%
  \includegraphics[clip,width=0.95\textwidth]{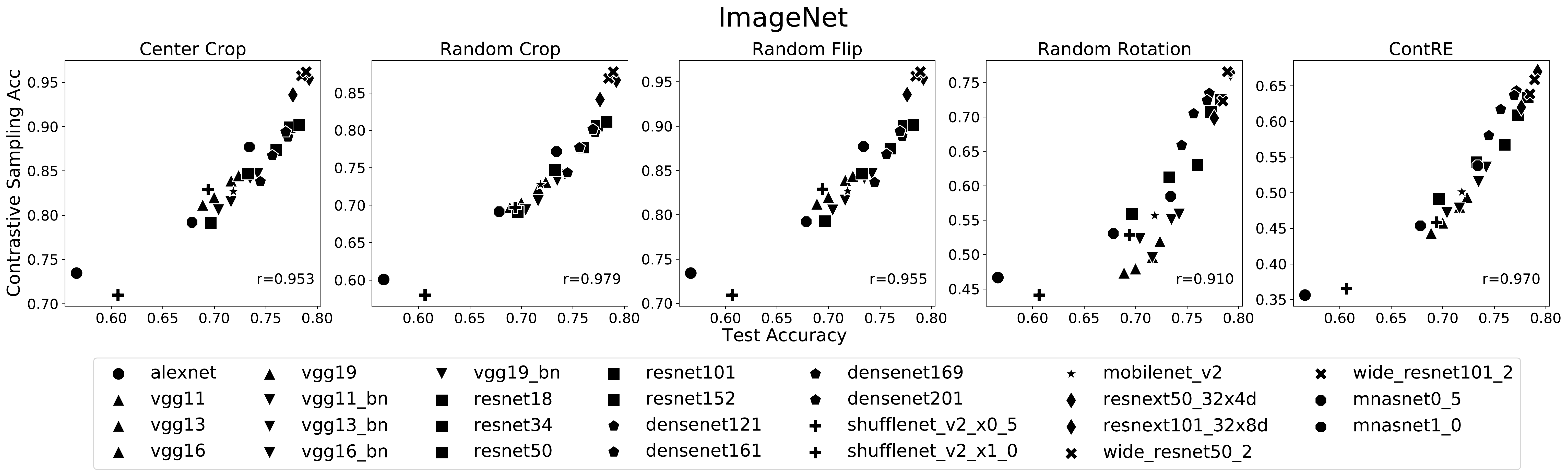}%
}
\caption{Spearman correlations between test accuracies and contrastive examples based accuracies on training data, using four trivial data augmentation techniques and our proposed approach \TheName{}, evaluated on CIFAR-10, CIFAR-100 and ImageNet. For DNN models on ImageNet, we use a single mark for the same family of networks due to the clarity and notation reasons.}

\label{fig:spearman-three-datasets}
\end{figure*}

In this section, we validate, through extensive experiments, the main claim that $\gL_{\bm{\theta}} (R(\mZ))$ is strongly correlated to $\gL_{\bm{\theta}} (\mZ_{test})$.
We conduct experiments with various networks on three open source benchmark datasets, and all results with Spearman correlations and partial correlations support our findings.
Furthermore, we demonstrate the applicability of \TheName{} across several practical settings, e.g., model selection across hyper-parameter choices, across RandAugment-Trained models, with small-size datasets or within a practical competition setting.
Finally, we provide analyses for investigating the reasons that \TheName{} works well as an estimator of DNN models' robustness and generalization performance.

\subsection{Experiment Setup}
We describe the experiment setups of datasets, networks and the methods for generating contrastive examples. 

\subsubsection{Datasets}
Three of the most popular benchmark datasets are considered: CIFAR-10, CIFAR-100 and ImageNet.
CIFAR datasets contain 60,000 tiny images of resolution 32$\times$32 while ImageNet includes over 1.2M natural images.
Conventional split of training and test sets are followed for CIFAR datasets, while the validation set of ImageNet is used as testing set here.

\subsubsection{Networks}
For evaluating our proposed framework \TheName{}, the most popular or the most powerful DNN architectures are considered: VGGNet \cite{simonyan2015very}, ResNet \cite{he2016deep}, Wide ResNet \cite{sergey2016wide}, DenseNet \cite{huang2017densely}, ResNeXt \cite{xie2017aggregated}, ShuffleNet \cite{zhang2018shufflenet, ma2018shufflenet}, MNasNet \cite{tan2019mnasnet}, MobileNet \cite{howard2017mobilenets, sandler2018mobilenetv2} and their variants.
We train these DNN models with standard data augmentation methods, i.e., random crop and random flip, and standard training hyper-parameters.
We also investigate the different data augmentation methods and different hyper-parameter settings for training DNN models, and the strong correlation between $\gL_{\bm{\theta}} (R(\mZ))$ and $\gL_{\bm{\theta}} (\mZ_{test})$ still holds.
Generally, models for CIFAR datasets are smaller than those for ImageNet but the architectures are shared among models for these datasets.
Moreover, many pretrained models on ImageNet are publicly available\footnote{\url{https://pytorch.org/vision/stable/models.html}}, which greatly accelerate our setup process.

\subsubsection{Contrastive Examples}
As introduced previously, RandAugment \cite{cubuk2020randaugment} is used in the \TheName{} framework to generate contrastive examples, where we take all the available transformation operations from the open-sourced implementation\footnote{\url{https://github.com/ildoonet/pytorch-randaugment}}.
We note RA\_N$X$\_M$Y$, where $X$ and $Y$ are the two hyper-parameters to configure in RandAugment: the number of sequential distortion operations and the magnitude of each distortion.
For example, RA\_N2\_M9 means that 2 operations will be chosen randomly from all the available operations and the distortion of magnitude 9 will be applied\footnote{The maximum magnitude is 30. Note that even the maximum magnitude does not transform the image into a total noise.}. 

\subsection{Experiment Results}

\begin{table}[t]
\caption{Correlations between the test accuracy and contrastive examples based accuracy on the training set, evaluated on three datasets CIFAR-10, CIFAR-100 and ImageNet with 14, 14 and 27 DNN models respectively.}
\centering
\begin{tabular}{@{}lrrr@{}}
\toprule
                & CIFAR-10 & CIFAR-100 & ImageNet \\ \midrule
Center Crop     & 0.702                                                          & 0.677                                                           & 0.953                                                          \\
Random Crop     & 0.722                                                          & 0.789                                                           & \textbf{0.979}                                                          \\
Random Flip     & 0.702                                                          & {\ul 0.790}                                                           & 0.955                                                          \\
Random Rotation & {\ul 0.938}                                                          & 0.653                                                           & 0.910                                                          \\
\TheName{} (ours)   & \textbf{0.965}                                                          & \textbf{0.974}                                                           & {\ul 0.970}                                                          \\ \bottomrule
\end{tabular}
\label{table:main-result}
\end{table}


Strong correlations between the generalization performance $\{\gL_{\bm{\theta}_i} (\mZ_{test}) \}_{i=1,2,...,m}$ and the accuracy on contrastive examples $\{\gL_{\bm{\theta}_i}(\mZ^s)\}_{i=1,2,...,m}$, are found across the three evaluated datasets, CIFAR-10, CIFAR-100 and ImageNet.
Here we present the detailed experimental results for these three datasets in comparison with trivial image transformations, such as center crop, random crop, random flip and random rotation.

As shown in Figure~\ref{fig:spearman-three-datasets} (top and middle), on the CIFAR datasets, most of the models have achieved close to 100\% accuracy on the training set for training samples transformed by crop or flip, which have been used during the training process.
For these transformations, the correlations are around 0.7.
Random rotation is not used during the training process, while the samples transformed by random rotation do not achieve stable correlations: a high correlation coefficient is obtained for CIFAR-10 while a relatively low value for CIFAR-100.
Meanwhile, our proposed approach \TheName{} consistently outperforms others by reaching the highest correlation coeeficients for both CIFAR datsets.

The results are different for models on ImageNet.
Due to the tremendous amount of samples and classes, ImageNet models with the best validation accuracies failed to perfectly fit the training data. As shown in Figure~\ref{fig:spearman-three-datasets} (bottom), training accuracies based on the listed operations all correlate well with test accuracies. Results given by random crop attains the first place with a correlation of 0.979, followed closely by \TheName's 0.970. 




The experimental results show that \TheName{} gets high correlation coefficients $Corr$ (Equation \ref{eq:corr}) of 0.969 on CIFAR-10 and CIFAR-100 and 0.970 on ImageNet, indicating that the grades of the models under the evaluation of \TheName{} are significantly correlated to the generalization performances of these models.
Note that this observation is hard to get from trivial approaches, including random crop, rotation, flip etc., because of randomness and instability. 
Considering these trivial approaches, ``Random Rotation'' works well on CIFAR-10 but yields a relatively low correlation coefficient on ImageNet. Similarly, ``Random Crop'' works well on ImageNet but not on CIFAR-10 or CIFAR-100.
However, across three tasks and datasets, our approach \TheName{} perfectly grades the models, probably thanks to the stability from the expectation over the randomness.
We thus reasonably conclude that \TheName{} might have the potential to estimate the generalization performance of models with the use of training data and the contrastive techniques.

\subsection{Ablation Studies}

For further showing the stability and effectiveness of \TheName{}, we carry out the experiments for ablation studies on the choices of two hyper-parameters in RandAugment when generating contrastive examples, and on the comparisons between contrastive examples from RandAugment and from single data transformations or the compositions of two transformations.

\subsubsection{Choices of $N$ and $M$}

In our framework \TheName{}, two hyper-parameters in RandAugment can be tuned: $N$ denotes the number of transformations that are to be applied on the original examples, and $M$ denotes the magnitude of distortions. Though they both represent the amount of distortions, their effects on \TheName{} are varied. As shown in \ref{table:randaugment-n-m}, we have tested all the compositions of $N = \{2, 3, 4\}$ and $M = \{4, 9, 15, 20\}$. A large ($M=15$) magnitude of each distortion leads to a higher correlation but an additional increase from 15 to 20 does not change much. The results match our previous observations, implying that a weak distortion on the samples cannot help to tell good classifiers from the bad ones. On the other hand, applying more transformations at the same time seems to hurt the performance, especially when $M$ is small. 
Therefore, our approach adapts a large value of $M=20$ and a smaller value of $N=2$, which is proved to be quite robust across various tasks and datasets.

\begin{table}[h]
\caption{Spearman correlations between test accuracy and contrastive examples based accuracy on the training set, with different choices of $N$ and $M$.}
\centering
\begin{tabular}{@{}lrrr@{}}
\toprule
                &$N$=2 & $N$=3 & $N$=4 \\ \midrule
$M$=4     & 0.837                & 0.763         & 0.503      \\
$M$=9     & 0.855             & 0.833        & 0.714        \\
$M$=15     & 0.974             & 0.974     & 0.960     \\
$M$=20 & 0.965    & 0.978   & 0.938    \\ \bottomrule
\end{tabular}
\label{table:randaugment-n-m}
\end{table}

\begin{figure}[t]
    \centering
    \includegraphics[width=\columnwidth]{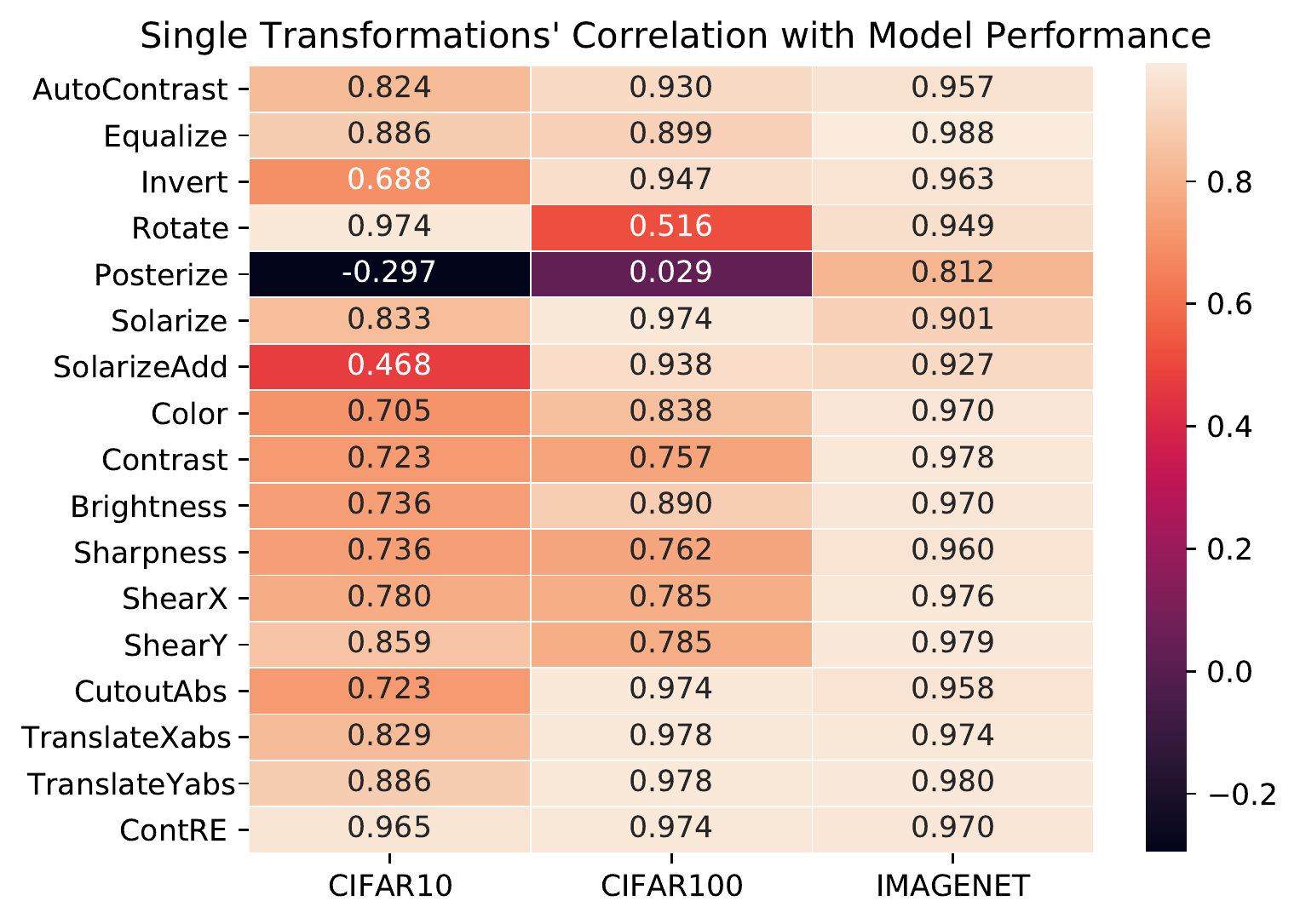}
    \caption{Spearman correlations between test accuracy and the accuracy on contrastive examples generated by single transformations and \TheName{}.}
    \label{fig:single-transform}
\end{figure}

\begin{figure}[t]
    \centering
    \subfloat[CIFAR-10]{\includegraphics[width=\columnwidth]{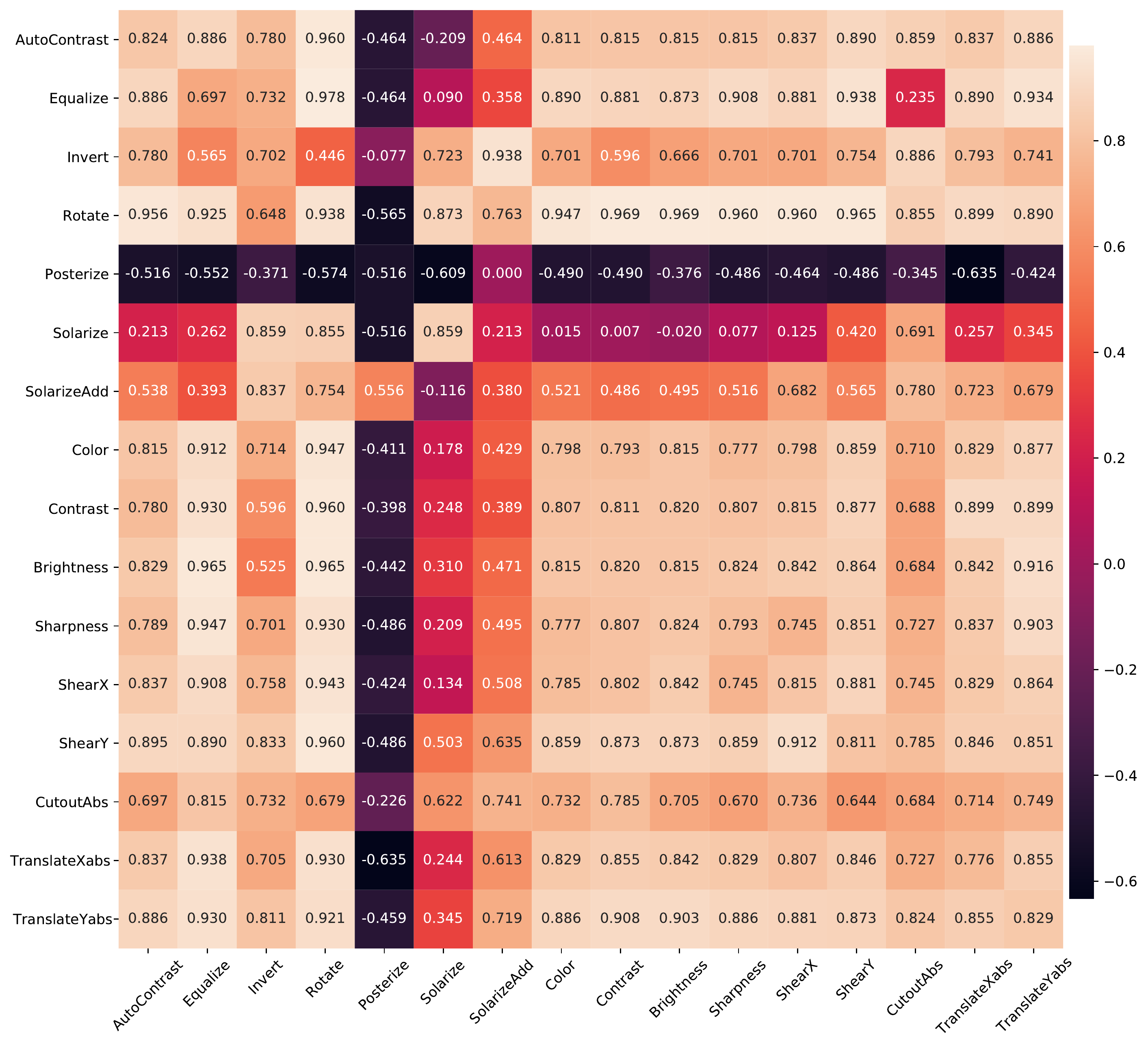}}
    
    \subfloat[CIFAR-100]{\includegraphics[width=\columnwidth]{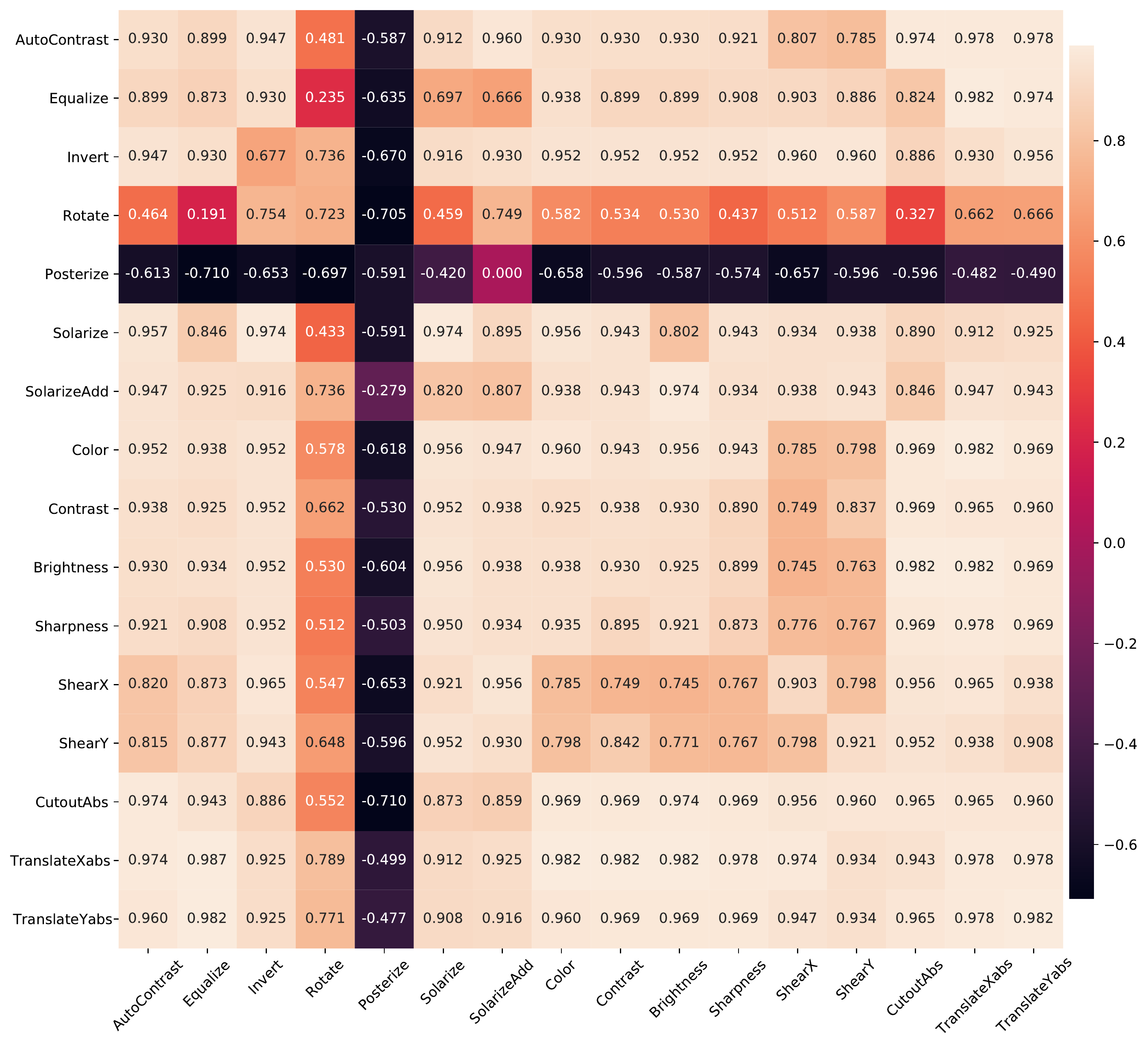}}
    
    \caption{Spearman correlations between test accuracy and the accuracy on contrastive examples generated by compositions of two transformations on CIFAR-10 and CIFAR-100.}
    \label{fig:compositions}
\end{figure}


\subsubsection{Single Transformations}
\TheName{} generalizes the idea of contrastive learning to the evaluations of DNN models' robustness and generalization performance.
However, is it really necessary to use a stochastic module for generating the contrastive examples for \TheName{}?
Is there any single transformation that could achieve the desired results?
We answer these questions through the comparison of the effects by \TheName{} and each single contrastive technique.

Following the same experimental setup, contrastive examples are generated under each single transformation with the same magnitude as \TheName{} and the experiments are performed using all DNN models considered previously.
The obtained Spearman correlation coefficients range from -0.297 to 0.988.
While a part of single transformations yield correlations higher than 0.9, especially for ImageNet-trained models, there is no single operation that performs well across the three datasets.
Therefore, choosing a best transformation depends on the dataset and it is difficult to find a single transformation that works well across different datasets. 
In contrast, \TheName{} introduces robustness and consistently achieves high performance through the expectation over the stochastic module and by combining several competent candidates.






\subsubsection{Compositions of Two Transformations}
To further prove the effectiveness and efficiency of \TheName{}, we conduct similar experiments on CIFAR datasets using contrastive examples that are generated by all the possible compositions of two transformations, obtain the accuracy on various contrastive examples and show the Spearman correlations with the testing accuracy in Figure~\ref{fig:compositions}.
Some observations are shared with the results from the comparison using single transformations, that any composition with ``Posterize'' does not work well for CIFAR datasets and that any composition with ``Rotate'' works well for CIFAR-10 but not for CIFAR-100.
A small part of compositions obtain higher correlations than \TheName{}.
We argue that searching the optimal composition demands much computation resource and the optimal choice varies from different datasets, while \TheName{} benefits the expectation over the stochastic module to consistently produce precise estimation results.
We will present in the following that \TheName{} is consistently applicable, efficient and effective in many practical complex situations.

\subsection{Evaluations for Applicability Analyses}
We have validated our proposed method \TheName{} on three popular datasets with standard training processes.
In this subsection, we carry out experiments with practical settings and demonstrate the applicability of \TheName{} in real-world situations.




\begin{figure*}[t]
    \includegraphics[width=0.85\textwidth]{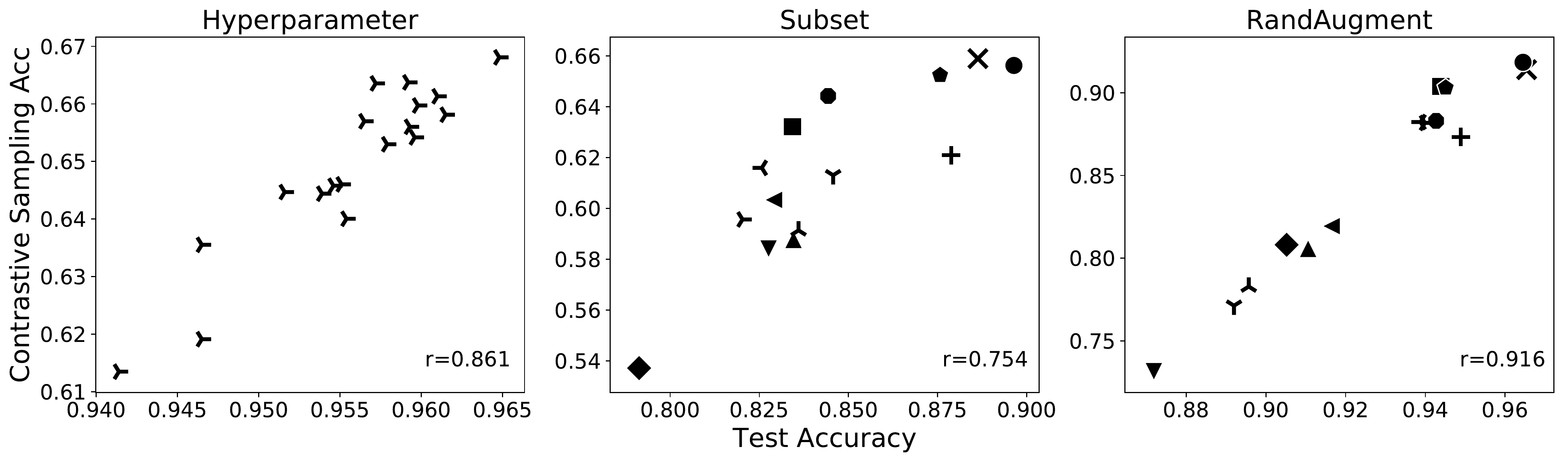}
    \caption{Spearman correlations between test accuracy and contrastive sampling accuracy on the training set for Wide-ResNet-26-10 with different hyperparameter choices (left); for all the models trained with 20\% training data of CIFAR10 (middle); for all the models trained with RandAugment on CIFAR10 (right).}
    \label{fig:secondary}
\end{figure*}

\begin{table}[t]
\caption{Spearman correlations between test accuracy and the accuracy on the contrastive examples, for models trained with different hyperparameter choices, trained with 20\% training data, and trained with RandAugment.}
\centering
\begin{tabular}{@{}lrrr@{}}
\toprule
                & Hyperparameter & Subset & RandAugment \\ \midrule
Center Crop     & 0.268                                                          & 0.465                                                           & 0.723                                                          \\
Random Crop     & 0.221                                                          & 0.535                                                           & 0.738                                                         \\
Random Flip     & 0.268                                                          & {\ul 0.583}                                                           & \textbf{0.978}                                                          \\
Random Rotation & {\ul 0682}                                                          & 0.055                                                           & {\ul 0.960}                                                          \\
\TheName{} (ours)   & \textbf{0.861}                                                          & \textbf{0.754}                                                           & 0.916                                                          \\ \bottomrule
\end{tabular}
\label{table:secondary-result}
\end{table}

\subsubsection{Across Hyper-parameter Choices}
A direct application of \TheName{} is to find the best model among models with the same architecture but trained with different hyper-parameters. We have trained 18 models, all of them is of the network structure of Wide-ResNet-26-10, on CIFAR-10 with various choices of batch size (32, 64, 128), learning rate (0.01, 0.05, 0.1) and weight decay (0.0001, 0.001). Based on the resulting 18 models, \TheName{} outperforms other approaches by achieving a correlation coefficient Corr of 0.859, see the results in Figure~\ref{fig:secondary} (left) and the comparison with other approaches in Table~\ref{table:secondary-result}. The results illustrate that, besides helping with selecting the best architecture, our approach can be helpful in the hyper-parameter-tuning stage.

\subsubsection{With Subset of Training Data}

In some practical settings, training data are very scarce while the experiments we have shown are using a large dataset with over 100 samples per class.
So we here would like to test the robustness of \TheName{} by reducing the training data size to small fraction (20\%) and training the DNN models. 
When provided with 20\% training data, models have limited information of the true data distributions and the generalization performances drop by 3-10 percent. 
Under such setting, \TheName{} outperforms trivial approaches by achieving a correlation coefficient of 0.754. The results indicate the potential of \TheName{} in predicting model performances when models are not trained with abundant data.

\subsubsection{Across RandAugment-Trained Models} 
Given \TheName{}'s high correlations between the generalization performance and the accuracy on contrastive examples of DNN models that are trained with standard data augmentation methods, we have observed that the ``Random Crop'' and ``Random Flip'' that are used for the standard training process do not provide desired results in our framework.
For over-parametric DNN models, including more data augmentation methods could usually improve the generalization performance.
We thus ask if the correlation obtained by \TheName{} mainly benefits from the image transformation methods that are not used during training and whether \TheName{} still performs well when the generated contrastive examples have already been used during training.

To answer the questions, we have trained the same set of DNN models on CIFAR-10 with replacing the standard data augmentation by the RandAugment module
while leaving other training settings unchanged.
Then, we repeat the processing of \TheName{} and get that \TheName{} still achieves a correlation higher than 0.9, with the results in Figure~\ref{fig:secondary} (right) and comparisons in Table~\ref{table:secondary-result}.
We note that ``Random Flip'' and ``Random Rotation'' get higher correlations than \TheName{}.
We argue that single transformations could get high correlations if they are not used during training, but the correlation coefficients largely decrease if they have been used during training, as shown in the previous results.
In contrast, \TheName{} combines a number of single transformations to generate contrastive examples and introduces the robustness from the expectation over the stochastic module.
Even with the models that have seen the contrastive examples during the training stage, \TheName{} can also successfully estimate the generalization performances.









\subsubsection{Generalization Gap Prediction~\cite{jiang2020neurips}}
\label{subsubsection:gap}


\TheName{} directly estimates the generalization performance instead of the generalization gap (the difference between the training accuracy and the generalization performance).
In the previous experiments, we mainly discussed the estimations of generalization performance in this work.
Meanwhile, under the general condition that all DNN models equally fit on the training set~\cite{jiang2020Fantastic}, \TheName{} is also capable of providing a good estimator of the gap.

To further demonstrate the feasibility of \TheName{}, we use the experiment settings of ``Predicting Generalization in Deep Learning Competition'' at NeurIPS 2020~\cite{jiang2020neurips} to evaluate \TheName{} on predicting the generalization gap of DNN models using the contrastive examples of the training set. 
The competition offers a large number of deep models trained with various hyper-parameters and DNN architectures trained on CIFAR-10 or SVHN, while the evaluator of the competition first estimates the generalization performance of every model using the proposed measure, then computes the mutual information score (the higher the better) between the proposed measures and the (observable) ground truth of generalization gaps. 

In the experiments, we propose to use the difference between the accuracy based on the original training samples and the one using \emph{contrastive examples} generated from the original training samples with \TheName{}.
For the comparison reason, we include several baseline measures in generalization gap predictors, including VC Dimension~\citep{vapnik2013nature}, Jacobian norm w.r.t intermediate layers~\citep{pgdl2020}, distance from the convergence point to initialization~\citep{nagarajan2019generalization}, and the sharpness of convergence point~\citep{jiang2019fantastic}, all of these baselines provide a theoretical bound for the estimated gap.
Table~\ref{tab:gengap_pred} presents the comparisons between the proposed measures and baselines. 
It shows that the proposed \TheName{} framework, with estimating the gap between the accuracy on original training samples and the one on contrastive examples, significantly outperforms the baseline methods.
Note that our proposed approach \TheName{} is an efficient and effective estimator for the generalization performance in practice, and that \TheName{} is a good estimator for the generalization gap in condition that the training accuracy be controlled on the same level.








\begin{figure*}[t]
    \centering
    \subfloat[Across Datasets]{\includegraphics[width=0.72\textwidth]{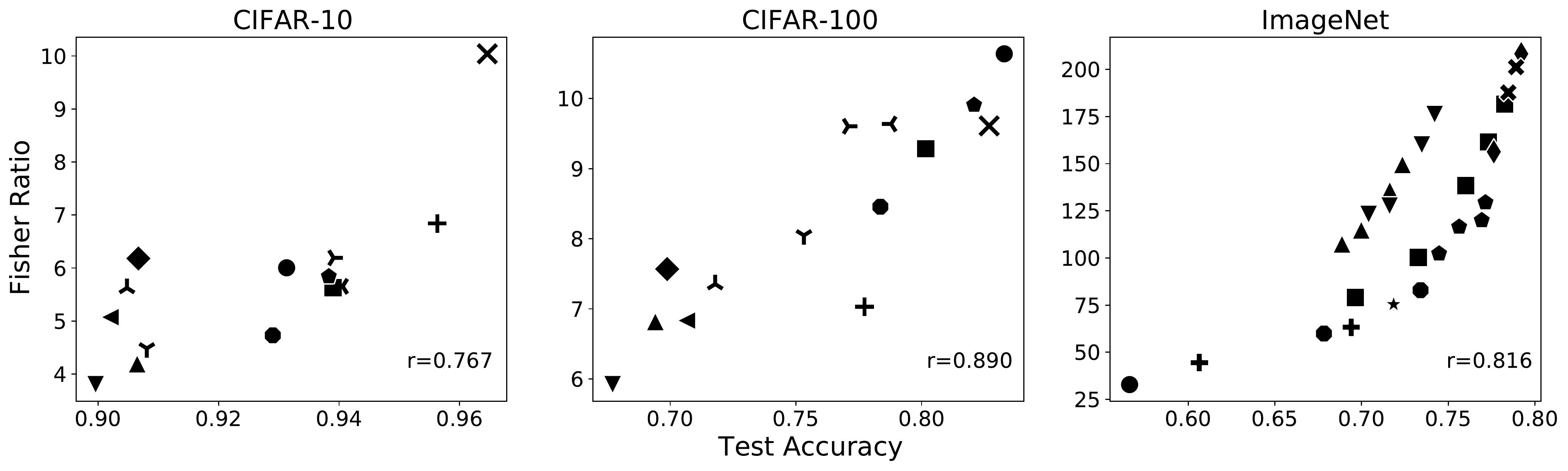}}
    \subfloat[Across Hyperparameter]{\includegraphics[width=0.23\textwidth]{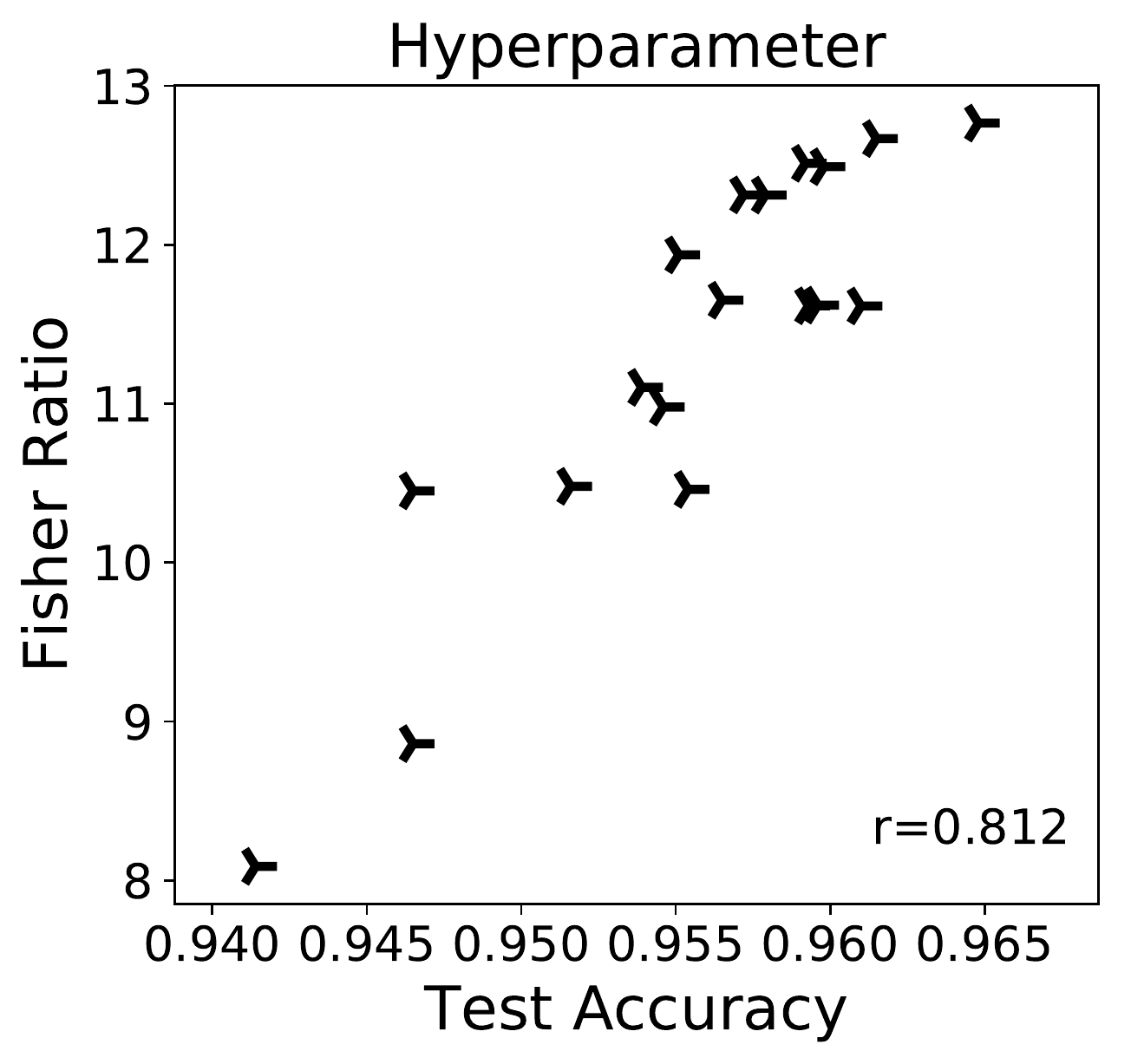}}
    
    \caption{Spearman correlations between test accuracy and Fisher ratios on contrastive examples.}
    \label{fig:fisher-ratios}
\end{figure*}

\begin{table}[t]
    \centering
    \caption{Mutual information scores of different methods to predict the generalization gap using CIFAR-10 and SHVN.}
    
    \begin{tabular}{lc}
        \toprule
        Methods	&	Prediction Score $\uparrow$ \\
        \midrule
        VC Dimension~\citep{vapnik2013nature}	&	0.020 	\\
        Jacobian norm w.r.t intermediate layers~\citep{pgdl2020}	&	2.061 	\\
        Distance to Initialization~\citep{nagarajan2019generalization}	&	4.921 	\\
        Sharpness of the convergence point~\citep{jiang2019fantastic}	&	10.667 	\\
        \TheName~(ours)	&	\textbf{13.531} 	\\
        \bottomrule
    \end{tabular}
    \label{tab:gengap_pred}
\end{table}

\subsection{Analyses}

We provide investigations from the partial correlations controlling the confounding variable and the feature-level correlations, to dissect the reasons that \TheName{} works for estimations of DNN models' robustness and generalization performance. 

\subsubsection{Partial Correlations}

\begin{table}[t]
\caption{Partial correlations between test accuracy and contrastive examples based accuracy on the training set while controlling training accuracy.}
\centering
\begin{tabular}{@{}lrrr@{}}
\toprule
                & CIFAR-10 & CIFAR-100 & ImageNet \\ \midrule
Random Crop     & 0.241   & 0.551   & {\ul 0.766}   \\
Random Flip     & N/A\footnotemark & {\ul 0.619}    & 0.228 \\
Random Rotation & {\ul 0.887}   & 0.190   & 0.566    \\
\TheName{} (ours)   & \textbf{0.938}    & \textbf{0.962}   & \textbf{0.774}    \\ \bottomrule
\end{tabular}
\label{table:partial-correlation-result}
\end{table}

\footnotetext{The correlation between accuracy of training examples after random flip and original training accuracy is 1, leading to division by 0 in Equation (\ref{eq:2}). Similar for ``Center Crop'' on the training set that is used as training accuracy.}






We observe that the training accuracy is well correlated to the testing accuracy on ImageNet, where the best model cannot fit the training set.
This shows that the training accuracy might be a good estimator of the generalization performance in some cases.
On the other hand, training accuracy is also related to the accuracy on contrastive examples because they are directly generated from the training set.
A question naturally follows: \textit{is the training accuracy the key factor for \TheName{} being a good estimator?}

To answer this question, we first define the notion of \textit{consistency} $\mC_{\bm{\theta}} (\mZ^s)$ as the difference between training accuracy and contrastive examples based accuracy on the training set $\mC_{\bm{\theta}} (\mZ^s) =\gL_{\bm{\theta}} (\mZ) - \gL_{\bm{\theta}} (\mZ^s)$.
The testing accuracy and the accuracy on contrastive examples contain a common variable, i.e., training accuracy $\gL_{\bm{\theta}} (\mZ)$, shown in the decomposition below:
\begin{equation}
    \left\{
         \begin{array}{ll}
         \gL_{\bm{\theta}} (\mZ^s)&= \gL_{\bm{\theta}} (\mZ) - \mC_{\bm{\theta}} (\mZ^s),  \\
         \gL_{\bm{\theta}} (\mZ_{test})&= \gL_{\bm{\theta}} (\mZ) - gap.
         \end{array}
    \right.
\end{equation}


With controlling the effect of this compound variable, training accuracy $\gL_{\bm{\theta}} (\mZ)$, we compute the partial correlation between the accuracy on contrastive examples $\gL_{\bm{\theta}} (\mZ^s)$ and the testing accuracy $\gL_{\bm{\theta}} (\mZ_{test})$.
Directly estimating the gap with the consistency fails on ImageNet for the reason that models fit to ImageNet training set in varying degrees.
This is not a practical issue for two reasons: (1) it will be less needed to estimate the gap if the generalization performance is strongly correlated to the training accuracy; (2) few datasets are in the scale of ImageNet.
Nevertheless, the results in Table~\ref{table:partial-correlation-result} show \TheName{} still gets relatively strong (partial) correlations on all the three datasets, while a moderate difference is observed from the comparison between CIFAR datasets and ImageNet.
We note that in the case of ImageNet, where the best model cannot perfectly fit the training set, the training accuracy is directly correlated to the testing accuracy and plays an important role for generalization performance estimation.
However, for CIFAR datasets, where all (or some) models get 100\% training accuracy, the training accuracy loses the predictability and the partial correlations controlling the training accuracy have no large difference compared to the standard correlations.
In summary, the training accuracy is not the key factor that \TheName{} works well for the generalization performance estimation in most cases.


\begin{table}[t]
\caption{Correlations between test accuracy and Fisher ratios of contrastive examples on the training set, evaluated on three datasets with 14, 14, 27 DNN models respectively.}
\centering
\begin{tabular}{@{}lrrr@{}}
\toprule
                & CIFAR-10 & CIFAR-100 & ImageNet \\ \midrule
Center Crop     & 0.495                                                          & 0.754                                                           & 0.662                                                          \\
Random Crop     & 0.455                                                          & {\ul 0.780}                                                           & 0.686                                                         \\
Random Flip     & 0.495                                                          & 0.754                                                           & 0.662                                                          \\
Random Rotation & {\ul 0.736}                                                          & 0.640                                                           & {\ul 0.774}                                                          \\
\TheName{} (ours)   & \textbf{0.767}                                                          & \textbf{0.890}                                                           & \textbf{0.816}                                                          \\ \bottomrule
\end{tabular}
\label{table:fisher-result}
\end{table}

\subsubsection{Feature-Level Correlations}

For complex DNN models, we further conduct analysis experiments by looking into the intermediate features of DNN models, computed on the original and contrastive examples. 
For these examples, we compute the Fisher ratios of features $f (\mZ; \bm{\theta})$ before the classifier for all models, and get the Spearman correlations with test accuracies. Note that for ImageNet and CIFAR-100 datasets, the number of data instances in each class is less than the number of dimensions of feature vectors when the DNN model, for instance DenseNet, is wide, leading to the non-invertibility of the within class matrix during the computation of Fisher ratio. 
We therefore perform the dimension reduction techniques on the feature vectors using singular value decomposition (SVD), reducing the number of dimensions to 64 or 512 for CIFAR-100 or ImageNet respectively, with most models containing over 95\% of the total variance.

The results in Figure~\ref{fig:fisher-ratios} (a) show that Fisher ratios from \TheName{} are highly correlated with the generalization performance of DNN models on each of the three datasets.
We also conduct experiments to compute the results on models with the same architecture but different hyper-parameters, shown in Figure~\ref{fig:fisher-ratios} (b). The conclusion also holds for models trained with different hyper-parameters.
The overall numerical comparison is shown in Table~\ref{table:fisher-result}.
It suggests that, among contrastive examples based on various operations, models that are able to generate better class-separated features from \TheName{} samples are more likely to have better generalization performance. Intuitively, Fisher ratio, which can be seen as a measure of the clustering quality, is highly correlated with the final classification accuracies. Therefore, the high correlation between \TheName{} accuracies and generalization performance inherently comes from the association between \TheName{} accuracies and its features' degree of separation, and that between Fisher ratio of features and model generalization performance.

\section{Conclusion and Discussion}

In this work, we build a framework \TheName{} to generate contrastive examples and measure the consistency of DNN models' behaviors with contrastive examples or training examples as input.
This consistency can be exploited to estimate the generalization performance of DNN models, endowed by the assumption that robust DNN models with good generalization performance tends to giving consistent features and predictions from the same image under varying data transformations.
We adopt RandAugment to generate contrastive examples and have practically assessed that the proposed \TheName{} is able to consistently estimate the generalization performance of various DNN models on three benchmark datasets.
Systematical ablation studies and thorough analyses have also been provided to demonstrate the versatility of \TheName{} in complex real-world situations and to dissect the reasons  that \TheName{} works well for the estimation.

We further discuss the potential limitations in the estimation of generalization performance by our proposed framework:
(1) The data transformations may spoil the data and lead to total random guess in the worst case.
We show that the transformations used by \TheName{} would not significantly affect the generalization estimation, by additionally reporting the correlations between transformed training and testing data, i.e., $\gL_{\bm{\theta}} (R(\mZ))$ and $\gL_{\bm{\theta}} (R(\mZ_{test}))$ (0.983, 0.903 and 0.989 for CIFAR10, CIFAR100 and ImageNet respectively), but for further extensions, transformations should be cautiously chosen.
(2) Theoretical analyses of local elasticity~\cite{bousquet2002stability,deng2020toward,he2019local} seems related to our approach, but theoretical links among contrastive examples, generalization performance and this notion are not available yet.
(3) Lack of appropriate data transformations, it would be difficult to extend to other data formats, such as texts, audios, graphs etc. Future investigation on the existing transformations and exploration on new ones would address this limitation.






\newpage
\bibliography{reference}
\bibliographystyle{ACM-Reference-Format}

\end{document}